\documentclass[letterpaper, 10pt, conference]{ieeeconf}
\IEEEoverridecommandlockouts                              
\usepackage{flushend}
\usepackage{algorithm}
\usepackage{algorithmic}

\usepackage{amsfonts}
\usepackage{url}

\usepackage{cite}

\ifx\pdfoutput\undefined
\usepackage{graphicx}
\else
\usepackage[pdftex]{graphicx}
\fi

\usepackage{stfloats} 
\usepackage{amsmath}  

\usepackage{color}

\renewcommand{\baselinestretch}{.95}

\title{\LARGE \bf
Sample-Based Planning with Volumes in Configuration Space
}

\author{Alexander Shkolnik and Russ Tedrake%
\thanks{A. Shkolnik is a research affiliate in the Computer Science and Artificial Intelligence Lab,
        MIT, 32 Vassar St., Cambridge, MA 02139, USA
        {\tt\small shkolnik@mit.edu}}%
\thanks{R. Tedrake is an Assistant Professor of EECS at the Computer Science and Artificial Intelligence Lab,
        MIT, 32 Vassar St., Cambridge, MA 02139, USA
        {\tt\small russt@mit.edu}}%
}

\begin{document}

\maketitle
\thispagestyle{empty}
\pagestyle{empty}

\begin{abstract}

A simple sample-based planning method is presented which approximates connected regions of free space with volumes in Configuration space instead of points. The algorithm produces very sparse trees compared to point-based planning approaches, yet it maintains probabilistic completeness guarantees.  The planner is shown to improve performance on a variety of planning problems, by focusing sampling on more challenging regions of a planning problem, including collision boundary areas such as narrow passages.

\end{abstract}

\section{Introduction}

Motion planning algorithms are used to solve challenging problems in robotics, assembly / disassembly tasks, and even drug discovery.  The field has made significant headway towards efficiently solving these P-Space hard problems.  An enabling development was the seminal idea of the Configuration Space \cite{Lozano-Perez83}, where the state of a robot is characterized by a single point, which is either in collision, $C_{obs}$, or is in free space, $C_{free}$.  A solution is then a trajectory through $C_{free}$.   More recently, the development of sample-based planning methods, such as the Probabilistic RoadMap  (PRM) \cite{Kavraki96}, and the Rapidly-exploring Random Tree (RRT) \cite{Kuffner00}, have used the configuration space to formulate planning as a search problem where the obstacle and robot geometry do not need to be explicitly considered.  These algorithms are widely used because they are computationally efficient and simple to implement, but performance suffers when there are narrow passages, or when the C-space is high dimensional. 

The PRM and RRT are tree based algorithms, where nodes in the tree correspond to feasible (collision-free) configurations, and edges represent feasible motions between these nodes.  In this paper, we utilize volumes to characterize regions of connected free-space instead of points.   The premise is that rather than building a tree of connected points in configuration space, where the points, and therefore the tree have zero volume, each node in the tree can instead be associated with a non-zero volume, representing a set of trivially reachable points from the parent node in the tree.  By considering a volumetric tree, significantly fewer nodes are able to characterize the connectivity of configuration space, therefore significantly improving search efficiency, while still preserving completeness guarantees.   This idea is developed in this paper in the context of an RRT, for single-query type problems.   

In many planning problems, there may be relatively easy regions to plan in, consisting of wide-open regions of configuration space, as well as intermittently challenging regions, consisting of narrow regions.  Such a problem can be characterized by the bug trap toy example, consisting of a room with a narrow tunnel  coming from within the room which offers the only path to escape (see Figure \ref{f:bugtrap-rrt}).  Planning collision free movements within the room is relatively easy, as is planning outside the room. The challenge then is finding a connected region of free space between these two areas.

\begin{figure}[t]
\centering
\includegraphics[width=2.5in, trim=1.2in 1in 1in .5in, clip]{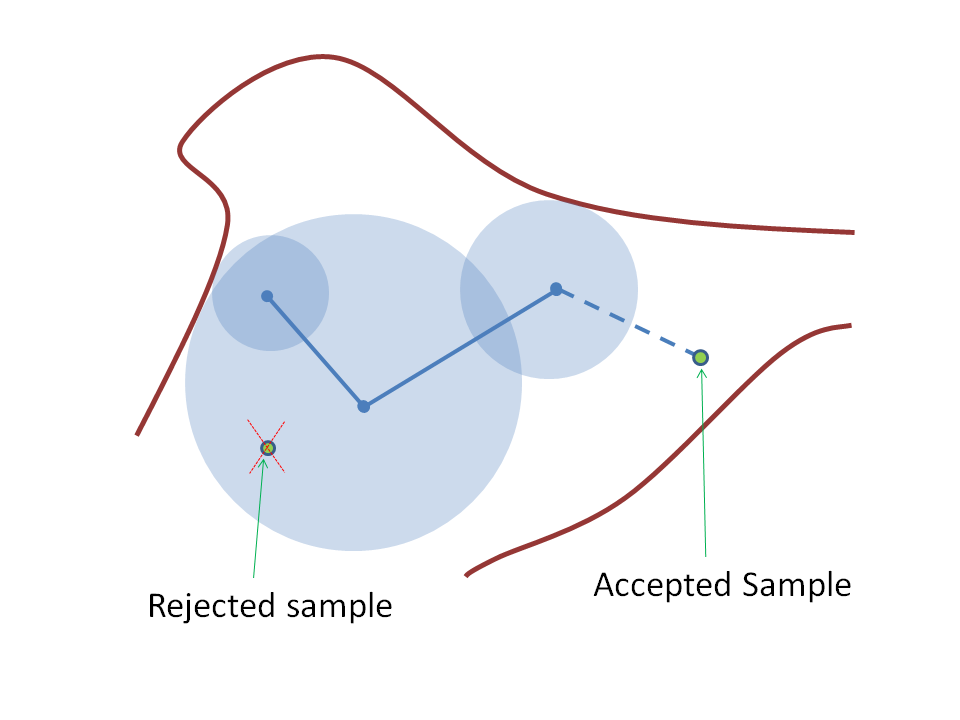} 

\caption{ Tree based planning with volumes. Edges are collision free paths between nodes; Nodes are volumes
characterizing free space. Each point in a node's volume has visibility to the node center. When growing the tree, samples within the tree volumes are rejected, and the tree is forced to expand to previously unexplored regions of free space.  }
 \label{f:volumes}
\end{figure}

The conventional RRT algorithm will not distinguish between difficult and easy areas of configuration space, and in fact will continuously add more and more samples within the room.  
One might visualize what it might be like to try to escape from a cluttered room while blind-folded, using only hands to feel around (just as the RRT is blind to explicit obstacle geometry).  One might quickly realize they are in a large room. Instead of searching for exits in random places (like the RRT would do), it would make sense to feel around the outer wall of the room until an exit is found.  A similar strategy can be employed by using volumetric trees for motion planning.  Parts of the tree will have nodes with large volumes, corresponding to areas of configuration space which are largely unencumbered by obstacles.  The planner can effectively neglect to search within these large regions of known free space, and instead focus the search on expanding into unexplored regions of configuration space.

In this paper, we build on the idea of planning with volumes.  A volume may be defined as the sublevel set of a distance (or cost) function.  When using Euclidean distance, the volume is simply a hypersphere.  We present a modified RRT which uses hyperspheres (balls) for planning, called the ``Ball Tree''.  Each node of the Ball Tree keeps track of the center and radius of the ball (the algorithm may also be generalized to other convex volumes). When building the tree, any samples that fall within the balls are discarded.  In this way, the connected regions of free space are approximated with the sum of center-to-center connected spherical volumes; any samples inside this volumetric tree will not lead to meaningful expansion, since paths within any ball to the center of the node is known, so these samples are safely rejected.  The sampling is then dynamically biased to encourage exploration of areas which the tree is currently unable to reach. 

 The remainder of the paper is organized as follows:  Section \ref{sec:background} provides a review of relevant background information.  Section \ref{sec:alg} describes the Ball Tree algorithm. An exact version of the Ball-Tree is presented first, which assumes perfect knowledge of distance from a point in free space to the closest collision surface. An inexact method is then described, which produces better results on more realistic problems where minimum-collision distance is not easily computable. This section also sketches a proof of completeness for both forms of the algorithm. In Section \ref{sec:results}, performance of the RRT is compared to the Ball-tree on four different planning problems. Finally, in Section \ref{sec:conclusion} we provide concluding remarks and discuss future directions. 


\section{Background and Related Work} \label{sec:background}
\label{s:background}

In this section, we review additional relevant material. As mentioned above, the algorithm is based on the RRT-Connect \cite{Kuffner00}, which the interested reader may wish to review.  
It is worth noting that a few other sample-based approaches have used volumes for planning. 
The ``Bubble Tree'' \cite{Brock01}, for example, builds tunnels in the workspace by growing a tree of overlapping spheres. 
When a sphere is inserted in the workspace the radius is bounded to the nearest collision surface.  
A priority queue of spheres is constructed based on distance to the goal. Using the priority 
queue to select a sphere, the surface of the sphere is randomly sampled, and a new sphere is 
inserted at the sampled surface point. Once a tunnel is found through the workspace, this 
information can be used to guide sampling for a robot in its full configuration space \cite{Rickert08}. This approach does not take advantage of the Voronoi bias which inherently improves search efficiency in an RRT, and also works well only for planning in the Workspace, where shortest collision distance is easily computed. Also related is the work by Morales et al. \cite{Morales07}, which utilizes bubbles of constant radius centered around some nodes in the tree to aggregate statistics and characterize the efficiency of sampling for a given random sample based planner. By grouping nearby nodes (those within a bubble), and measuring the ratio of bubbles to nodes, one can tune sampling parameters, and also spot problem areas where repeated growth attempts do not result in successful expansions. 

Another work which is conceptually closely related is the RESAMPL algorithm \cite{Rodriguez08}, which groups regions of samples by proximity (the regions could be spheres), and focuses sampling on select regions. Samples in configuration space are first grouped into regions. The regions are then characterized by their entropy as either being low entropy, where most samples are $C_{free}$ or $C_{obs}$, or high entropy regions, which contain surface regions, or narrow passages. A set of adjacent regions which are likely to contain a potential solution path can be explored.  In doing so, sampling can be adjusted to encourage exploration of high entropy areas as needed, and discourage sampling of low entropy areas which are more or less trivial (or impossible) to plan through.  The algorithm requires tuning for choosing how to best define regions, and also requires extensive sampling to appropriately classify regions containing very narrow passages. 

	In addition to these bubble and region based approaches, other approaches in the literature use workspace information to guide the growth of an RRT or PRM in the full configuration space (see for example \cite{Diankov08, Zucker08, Holleman00}).  The work presented in this paper can be used as an alternative method for very quickly and efficiently exploring 2D and 3D workspaces, from which more complete planning can take place, or the Ball Tree can be used to directly search in the higher dimensional configuration space.  Additionally, the Inexact Ball Tree does not require shortest collision distance information like the Bubble Tree. 

	Other related work comes from ideas in feedback planning. For an excellent review of motion planning, including planning with feedback controllers, see \cite{LaValle06}. Local feedback controllers have basins of attraction, which direct the robot from a state in the basin to some other bounded region of state space. The idea of funnels is to create local feedback controllers where any point in the controllers basin (funnel) gets directed into an adjacent funnel. A number of funnels can then be combined so their basins cover the state space with a feedback policy. 
	  A sample-based algorithm which uses spheres as funnels is described in \cite[p. 413]{LaValle06}.  First, a ``cover'' is created, which tries to fill a space with spheres. As samples are drawn, if they lie in free space, and are not contained within a sphere, then a new sphere is added centered at the sample. Like the Bubble Tree, spheres are added so they are touching a collision surface, requiring distance information.  
	  The process continues until substantial coverage is achieved. Paths can be found by considering each sphere as a potential field leading into overlapping neighboring balls. The approach is feasible for planning in 2D or 3D workspaces, or simple configuration spaces where shortest distance to obstacles can be efficiently computed. For planning problems with very narrow passages, for example the Alpha Puzzle 1.0 \cite{Yamrom}, the probability of sampling inside the passage is very low, which would inhibit this type of planner. 
	
	Building on the feedback idea is the LQR-tree \cite{Tedrake10}, designed to combine local feedback planners, based on time-varying LQR solutions, with global planning methods such as the RRT. A tree of local planners is assembled, grown backwards from the goal, where a conservative basin of attraction is estimated for each controller. Sampling continues until a substantial portion of state space is covered with basins, resulting in a feedback policy.   A simulation-based LQR-tree \cite{Reist10} was also developed, which verifies the size of the basins of attraction by simulation, instead of explicitly computing basins.  	The work in this paper builds on these results to try to build trees which cover the space as efficiently as possible. 
	
	The Ball Tree is also loosely based on the Reachability-Guided RRT (RG-RRT) \cite{Shkolnik09a}, which was developed for single-query planning problems in dynamic systems. When there are dynamics constraints, particularly nonholonomic and underactuated dynamics constraints, then standard planning with an RRT is hindered because the Euclidean distance metric fails to describe an appropriate distance for expanding nodes. One can also view the problem as if each node in the tree itself lies in a narrow tunnel (outlined by the constraints on the dynamics), so planning in state-space with dynamics magnifies the narrow-tunnel planning problems faced by most sampling based planners. The RG-RRT algorithm samples actions for each node in the tree, and stores the resulting states as "reachable" leaf nodes in the tree. When samples are drawn, samples which are closer to the tree nodes are rejected, and only samples close to reachable nodes are used to try to expand the tree. In this way, many samples are rejected, and the tree is only grown when it is likely that the expansion attempt will succeed, similar to the dynamic domain sampling method introduced in \cite{Yershova05} and \cite{Jaillet05}.  By considering the reachable set for kinematic type planning problems (where velocities and accelerations do not need to be considered), one can see that a node in the tree can move in any direction, so it makes sense to use the Euclidean distance metric, and the local reachable set can be approximated as a sphere. These ideas directly lead to the work in this paper, though the method for approximating and pruning the "reachable" sets in this work is different.  By changing the sampling distribution, the Dynamic Domain RRT encourages sampling near the difficult areas for expansion, and also encourages refining the inside nodes of the tree. In this work, we use volumes to discourage sampling the inside of the tree, unless there is reason to do so. This results in very sparse tree, where sampling is focused mainly on the nodes of the tree which are having difficulty growing into nearby open regions of configuration space.

\section{Algorithm} \label{sec:alg}
\label{s:algorithm}

\begin{figure*}[t]
\centering
\includegraphics[width=5in, trim=0in 2.6in 0in 1.1in, clip]{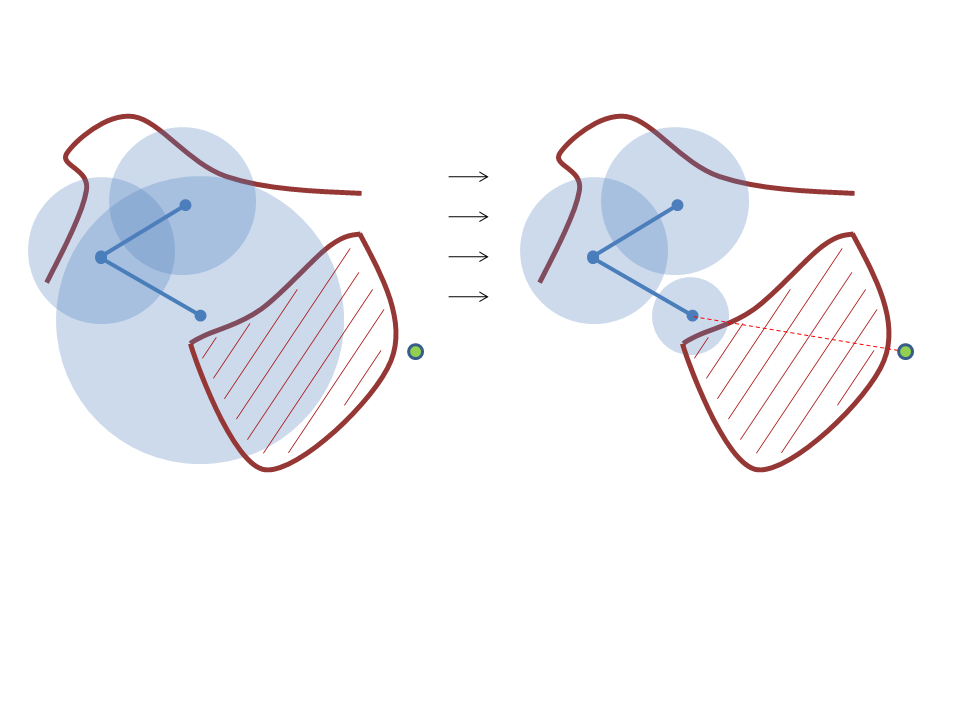} 

\caption{ Inexact version of Ball Tree:  tree volumes may overlap with obstacles. All edges are verified
to be collision free. When expanding to a new node (green point), if a collision is encountered, the volume is trimmed to reflect the collision distance information (trimmed volume shown on right).   }
 \label{f:trimming}
\end{figure*}

We present two variations on the Ball Tree algorithm.  In the first version, we assume that the largest collision-free hypersphere around a point can be computed efficiently and exactly.  Then we remove that assumption and instead use the collision checking on the random samples used in the algorithm to estimate the volumes.   Surprisingly, this second ``inexact'' algorithm often outperforms the ``exact'' algorithm in terms of computation time and/or number of nodes.  Both versions of the algorithm are probabilistically complete. 

\subsection{Exact Ball Tree}

We begin by exploring the Exact Ball Tree Connect algorithm, as described in Algorithms \ref{alg:ball-tree} -  \ref{alg:extend-ball-tree}.  The algorithm is similar in structure to RRT-Connect.  Two trees are grown iteratively, one from the start node, and one from the goal node.   For each node in the tree, v, we store the volume position, v.x, and a radius, v.r, which is the shortest distance to any obstacle from v.x.   A rejection step is introduced into the sampling step, \ref{alg:ball-tree} Line 5, whereby random samples which are contained inside any volume affiliated with either tree are rejected, and a new sample is drawn.  Aside from rejecting samples, the BALL-TREE algorithm and the CONNECT algorithm are identical to the RRT-CONNECT in \cite{Kuffner00}.  The EXACT-EXTEND is also similar to the classic RRT. However, when choosing an expansion node, the Ball-Tree refers to the Nearest-Volume, rather than the nearest point. Since the node-volumes are defined as a sublevel set for a given distance metric, the distance from the node-center to the volume-surface is constant for any point on the surface, and this distance can be subtracted when comparing distances from a sample to nodes of the tree. Using Euclidean distance, with hyper-spherical volumes, the Nearest-Volume function simply returns the node $i$ which minimizes  $||x_{rand} - v_i.x||_2 - v_i.r$. Additionally, the EXACT-EXTEND algorithm is tasked with computing and storing the shortest distance to obstacles in Line 4.

The exact ball tree is conceptually very simple, and allows one to easily visualize the idea of planning with volumes.  The algorithm is probabilistically complete, as shown in the sketch in the next section. However, for many problems, it can be difficult to find the minimum distance to collisions in arbitrary configuration spaces. A distance measure requires explicit consideration of the geometry, which negates some of the nice properties of the RRT which makes it fast to begin with.

\begin{algorithm}[H]
	\caption{BALL-TREE ($x_0$) }
	\label{alg:ball-tree}
	\begin{algorithmic}[1]
		\STATE $T_a.init(x_0)$
		\STATE $T_b.init(x_{goal})$
		\WHILE{\TRUE}
		    \STATE $x_{s} \Leftarrow$ RANDOM-SAMPLE() 
		    \IF{INSIDE($x_s$,$T_a$) \textbf{or} INSIDE($x_s$,$T_b$)}
				\STATE \textbf{continue}
                    \ENDIF
			\STATE $[v_{new},status] \Leftarrow$ EXACT-EXTEND$(T_a,x_{s})$
			\IF { $v_{new}$ \textbf{and} CONNECT$(T_b, v_{new}.x) 
                          == Reached$ }
				\RETURN PATH$ (T_a, T_b) $
			\ENDIF
			\STATE SWAP$(T_a,T_b)$
		\ENDWHILE
	\end{algorithmic}
      \end{algorithm}

\begin{algorithm}[H]
	\caption{CONNECT ($T, x$)}
	\label{alg:connect-ball-tree}
	\begin{algorithmic}[1]
		\REPEAT
		\STATE $[v_{new},status] \leftarrow$ EXTEND$(T,x)$
		\UNTIL{  $status \neq Advanced$ }
		\RETURN $status$
	\end{algorithmic}
\end{algorithm}

\begin{algorithm}[H]
	\caption{EXACT-EXTEND ($T, x$) }
	\label{alg:extend-ball-tree}
	\begin{algorithmic}[1]
		\STATE { $[v_{near}, mindist] \Leftarrow$  NEAREST-VOLUME($x$,T) }
		\STATE {$ \left[ u_{new}, v_{new}.x \right]  \Leftarrow $ NEW-STATE($v_{near}.x, x$) }
		\IF { VALID($u_{new}, v_{new}.x$) }
			\STATE $v_{new}.r$ = DIST-CLOSEST-OBS($v_{new}.x$)
			\STATE T.add-node($x_{near}, v_{new}, u_{new}$)
			\IF { $v_{new}.x == x$ }
					\RETURN [$v_{new},Reached$]
			\ELSE
					\RETURN [$v_{new},Advanced$]
			\ENDIF
		\ENDIF
		\RETURN [NULL,$Trapped$]
	\end{algorithmic}
\end{algorithm}

\subsection {Inexact Ball Tree}

Using similar principles of storing volumes in the nodes of the tree, and rejecting samples that land in the volumes, we can define an ``inexact'' version of the Ball Tree, which does not require a minimum collision distance computation. In this case, the EXTEND operation is modified, as shown in Algorithm \ref{alg:inexact-extend-ball-tree} and \ref{alg:check-overlap}. The idea is to initialize a volume with a large radius, $R_0$. This radius may even be infinite, though generally a more reasonable value may be more efficient. This radius is then trimmed when collisions are encountered during the NEW-STATE operation.  The NEW-STATE function attempts to connect the nearest tree node with a non-rejected sample, which lies outside the tree volumes.  If a collision is found, the radius of the tree node is reduced or ``trimmed'' to the lesser of the current radius or the estimated collision distance, as calculated within the NEW-STATE function.  In the results presented below, with the exception of the Alpha Puzzle, the NEW-STATE function is implemented by linearly interpolating the action, and checking each point for collision. If a collision is found, the ball radius is trimmed to the distance of the closest collision point found by NEW-STATE().

If the EXTEND operation succeeds, a new node is added to the tree containing a volume, with radius $R_0$. The CHECK-OVERLAP function is introduced to address overlapping volumes between trees. If the volume of the newly added node overlaps with some volume in $T_2$, then we attempt to connect the two trees. This step will either succeed, in which case we have succeeded in finding a path from start to goal, or the step will fail, in which case at least one or both radii are trimmed. This step is repeated until there are no more overlapping volumes found between the new node and $T_2$. This step ensures that trees do not overlap unless a connection between them is validated, a necessary step to ensure the completeness of the algorithm.  Note, that when the trees are initialized, the CHECK-OVERLAP can be performed. This forces the start nodes of the trees to have a reasonable radii, and ensures that regions of free space can still be sampled.


\subsection {Probabilistic Completeness of the Exact Ball Tree method}

Suppose we have a connected tree, T, as defined by the exact ball tree algorithm. Each node in the tree consists of a convex volume in $C_{free}$, with center, x, and convex volume definition V. The edges in the tree are collision free paths between node centers.  Because a node's volume is convex, for any point on the surface of (or within) this volume, there is a straight-line path in free space to the center, and we can therefore define a path from the root of the tree to any point lying on the surface of the tree volumes.  Regions of the surface can be categorized in one of three ways:  1) region or point which is in contact with $C_{obs}$;  2) regions of the surface which are contained within some other volume of the tree; and 3) the remainder of the surface, called free-surface, which is adjacent to $C_{free}$ and not contained within any tree node volume.   If we pick any point within the free-surface, then there exists some hypersphere, H,  centered around this point, with an arbitrarily small radius, $\epsilon > 0$, for which this hypersphere is entirely in free space. The volume defined by (H - tree-volume) is non-zero, therefore there is a strictly positive probability of sampling near any point of the free-surface of the tree. Furthermore, the distance metric is defined with respect to the node affiliated with this surface, so that the distance to a given node is zero along its surface.  
Therefore, we can guarantee that for any node in the tree which is exposed to free space (e.g. where part of the surface is not in collision, and also not contained inside another ball), there is a non-zero probability of extending that ball into its adjacent free space region. Furthermore, there is a non-zero probability of extending into any regions of free space which are $\epsilon$-close to the tree. 
Therefore, as the number of samples goes to infinity, the probability of covering the entire connected region of free space with balls converges to 1, implying that the exact Ball-tree algorithm is probabilistically complete.

\subsection {Probabilistic Completeness of the INEXACT Ball Tree method }

The proof is constructed in a similar manner to the one above. The difference is that in the inexact version of the algorithm, node volumes may overlap with regions of $C_{obs}$.   As noted above, a convex volume which is not in collision has straight-path visibility from regions adjacent to the surface of this volume to the center of the volume. In the case where volumes overlap with obstacles, we break down the surface into 4 categories, including: 1) surface regions adjacent or inside of collisions; 2) regions inside other node volumes;   3) regions adjacent to free-space, with straight-line collision-free connectivity to the center of the node; and 4) regions which are adjacent to free-space, but which do not have straight-line connectivity to the center node.  By the same logic that there is a non-zero probability of extending a node into any adjacent free-space,  we can say there is a non-zero probability of attempting to extend a node into regions of free space for which there is no straight-line path. In this case, the volume of the ball will be trimmed so that less of the ball is in collision. Therefore, with probability 1, any ball which contains surface regions of type (4) will be trimmed until this surface type does not exist.  When these regions are reduced to null, the proof of completeness is the same as for the Exact Ball tree case.

\begin{algorithm}[]
	\caption{INEXACT-EXTEND ($T, x, T_2$) }
	\label{alg:inexact-extend-ball-tree}
	\begin{algorithmic}[1]
		\STATE { $[v_{near}, mindist] \Leftarrow$  NEAREST-VOLUME($x$,T)  }
		\STATE {$ \left[ u_{new}, v_{new}.x \right]  \Leftarrow $ NEW-STATE($v_{near}.x, x$) }

		\IF { VALID($u_{new}, v_{new}.x$)  }
				\STATE $v_{new}.r = r_0$
				\STATE T.add-node($v_{near}, v_{new}, u_{new}$)
				\IF { CHECK-OVERLAP($v_{new}, T_2$) == Plan-found }
					\RETURN Plan-found
				\ENDIF
				\IF { $v_{new}.x == x$ }
						\RETURN [$v_{new}$, Reached]
				\ELSE
						\RETURN [$v_{new}$, Advanced]
				\ENDIF
		\ELSE
				\STATE {Trim $v_{near}.r$} \COMMENT{use collision information from NEW-STATE to trim the ball radius, if a collision is detected within the boundary of the ball.}
		\ENDIF
		\RETURN [NULL, Trapped]
	\end{algorithmic}
\end{algorithm}

\begin{algorithm}[]
	\caption{ CHECK-OVERLAP($v_{new}, T_2$) }
	\label{alg:check-overlap}
	\begin{algorithmic}[1]
				
				\STATE Overlap = \TRUE
				
				\WHILE {Overlap}
					\STATE 
\begin{flushleft}
					$\left[ v_{close}, mindist \right] $ = NEAREST-VOLUME($v_{new}.x, T_2$)
\end{flushleft}

					\IF { $mindist < 0$ }
						\STATE {$ \left[ u_{t}, x_{t} \right]  \Leftarrow $ NEW-STATE($v_{new}.x, v_{close}.x$) }
						\IF { VALID($u_{t}, x_{t}$)  }
							\IF {$x_{t} == v_{close}.x$ }
								\RETURN Plan-found
							\ENDIF
						\ELSE
						\STATE {Trim $v_{new}.r$ and $v_{close}.r$} 
						\COMMENT{use collision information from NEW-STATE to trim the ball radius if a collision is detected within the boundary of either ball.}
						\ENDIF
					\ELSE
						\STATE Overlap = \FALSE
					\ENDIF
				\ENDWHILE
				
				\RETURN No-overlap
				
	\end{algorithmic}
\end{algorithm}

\subsection { $\delta$-Ball Tree }

A simple but powerful modification of the inexact ball tree allows balls to penetrate obstacles by a small slack factor, delta.  In this case, when trimming, the ball radius is set to the minimal collision distance plus $\delta$.  If a ball has been trimmed to the minimum radius, then further attempts to trim it down will be unsuccessful. This offers a trade-off. On the one hand, the minimum ball radius introduces bounds into the planning algorithm, guaranteeing a solution to be found within a finite number of node expansions. Consider that a ball touches a surface at only a point, so to completely fill a prismatic volume with overlapping spheres, infinite spheres are required. On the other hand, if the spheres are allow to penetrate by some distance into the walls of the prism, then a finite number of balls can completely fill the prism. Of course, this slack factor destroys the basic premise of the probabilistic completeness proof given above. However, it can be shown that the algorithm remains complete if there exists a solution path from start to goal, for which the closest distance from any point on the path to any obstacle is greater than $\delta$.  In practice, this modification can significantly improve search time, even for small values of $\delta$.

\section {Results}  \label{sec:results}
\label{s:performance}

\begin{figure*}
\centering
\includegraphics[width=2.8in, trim=3in 1in 2.5in .5in, clip]{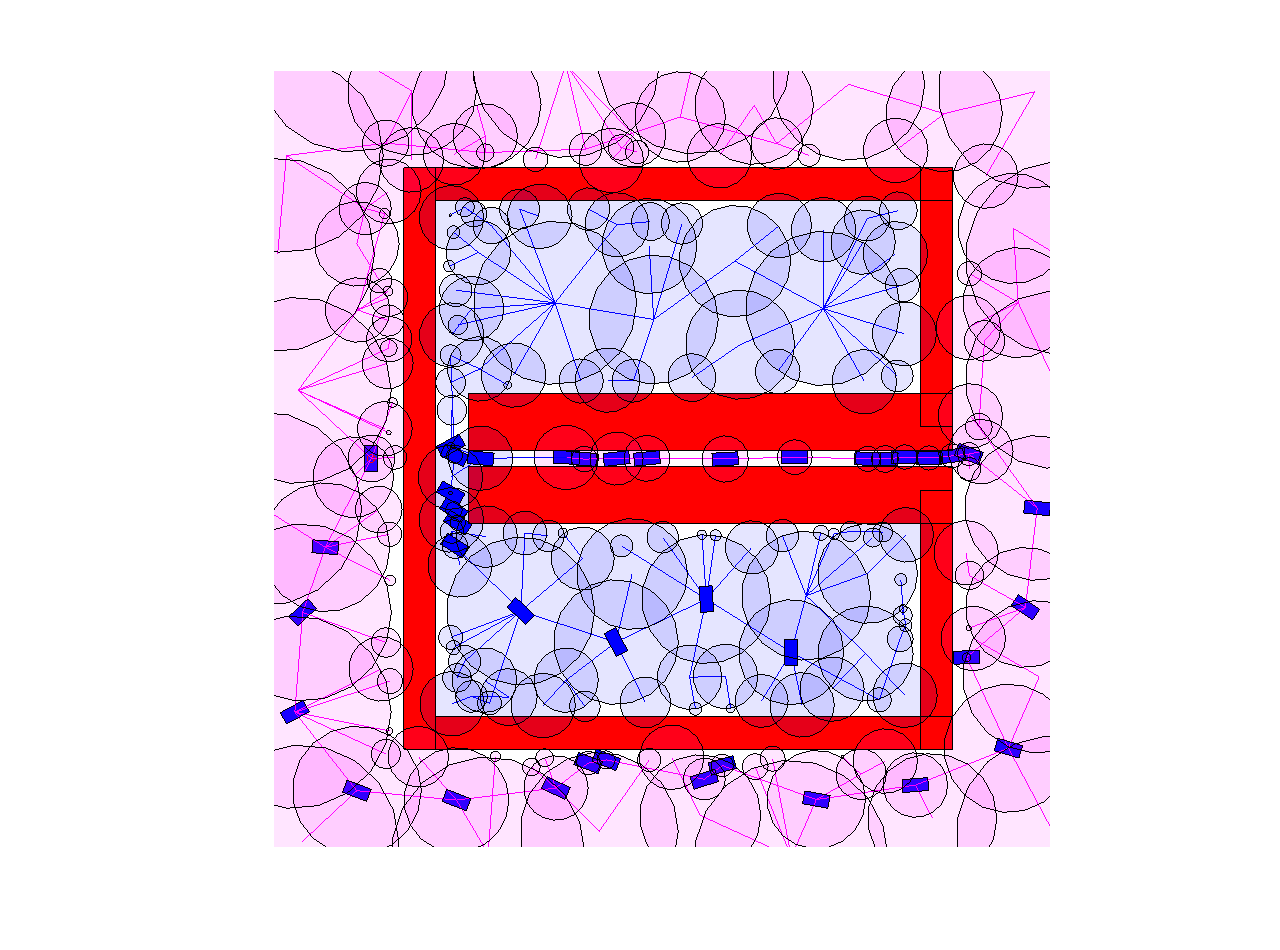} 
\hspace{.4in}
\includegraphics[width=2.8in, trim=3in 1in 2.5in .5in, clip]{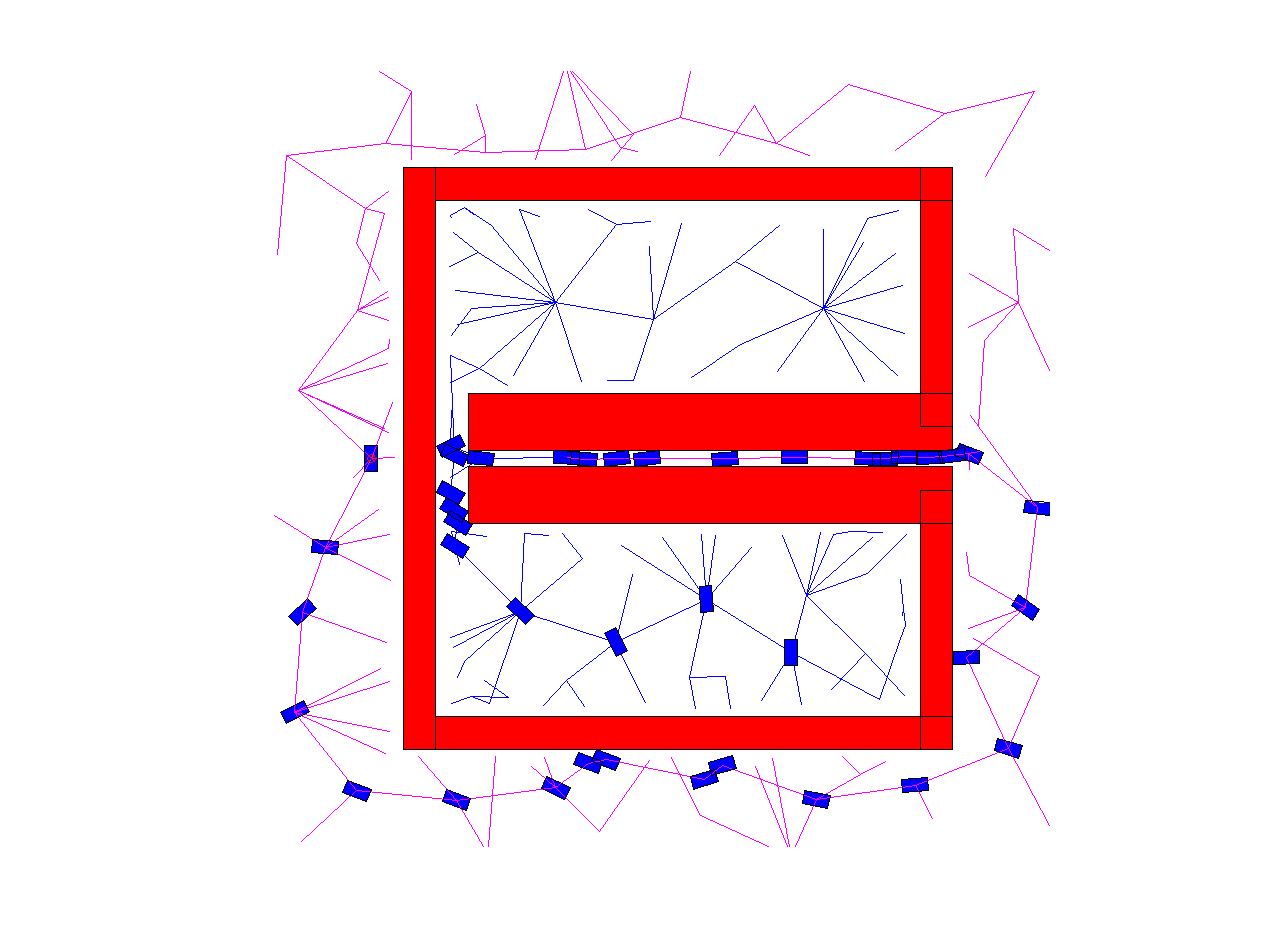} \\
\caption{ Left: Ball Tree solution of bug trap. Right: Showing same Ball-Tree, but just the node centers. }
 \label{f:bugtrap-balls}
 \vspace{-10pt}
\end{figure*}

\begin{figure}[h!]
\centering
\includegraphics[width=2.8in, trim=3in 1in 2.5in .5in, clip]{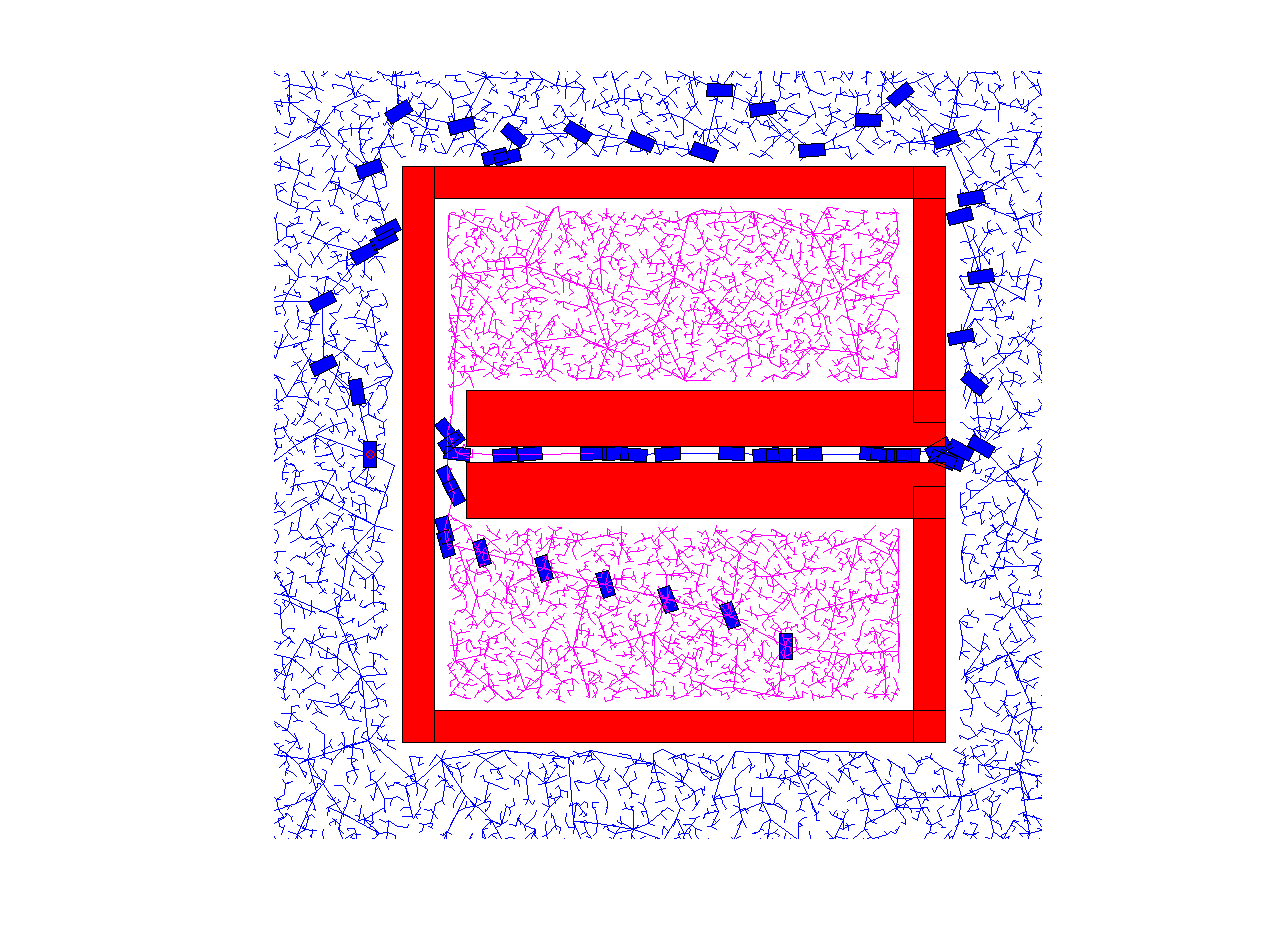} 
\caption{ RRT solution of bug trap.  }
 \label{f:bugtrap-rrt}
 \vspace{-10pt}
\end{figure}

\begin{figure*}[!ht]
\includegraphics[width=2.2in, trim=3.5in 1in 3in .5in, clip]{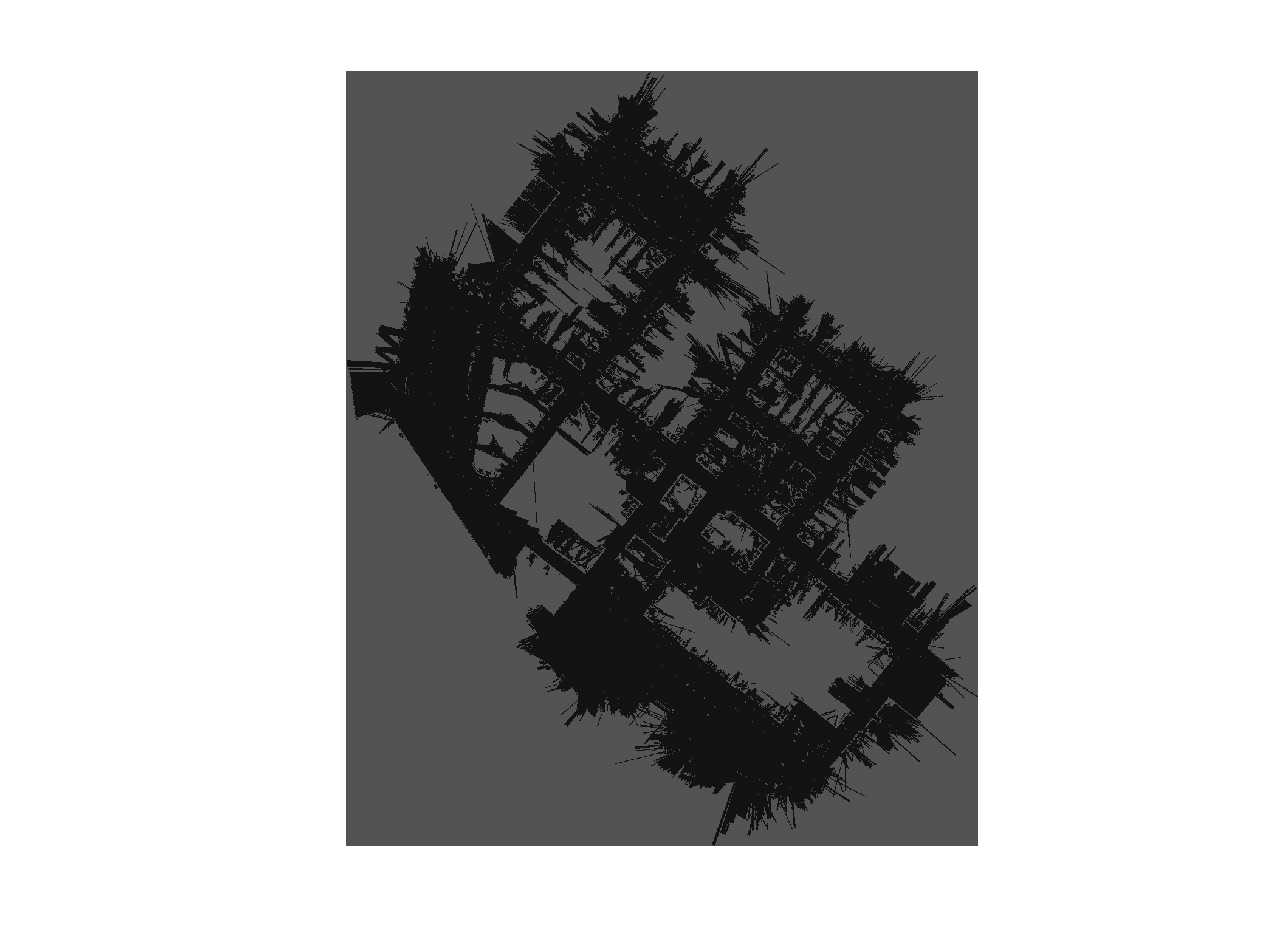}
\includegraphics[width=2.2in, trim=3.5in 1in 3in .5in, clip]{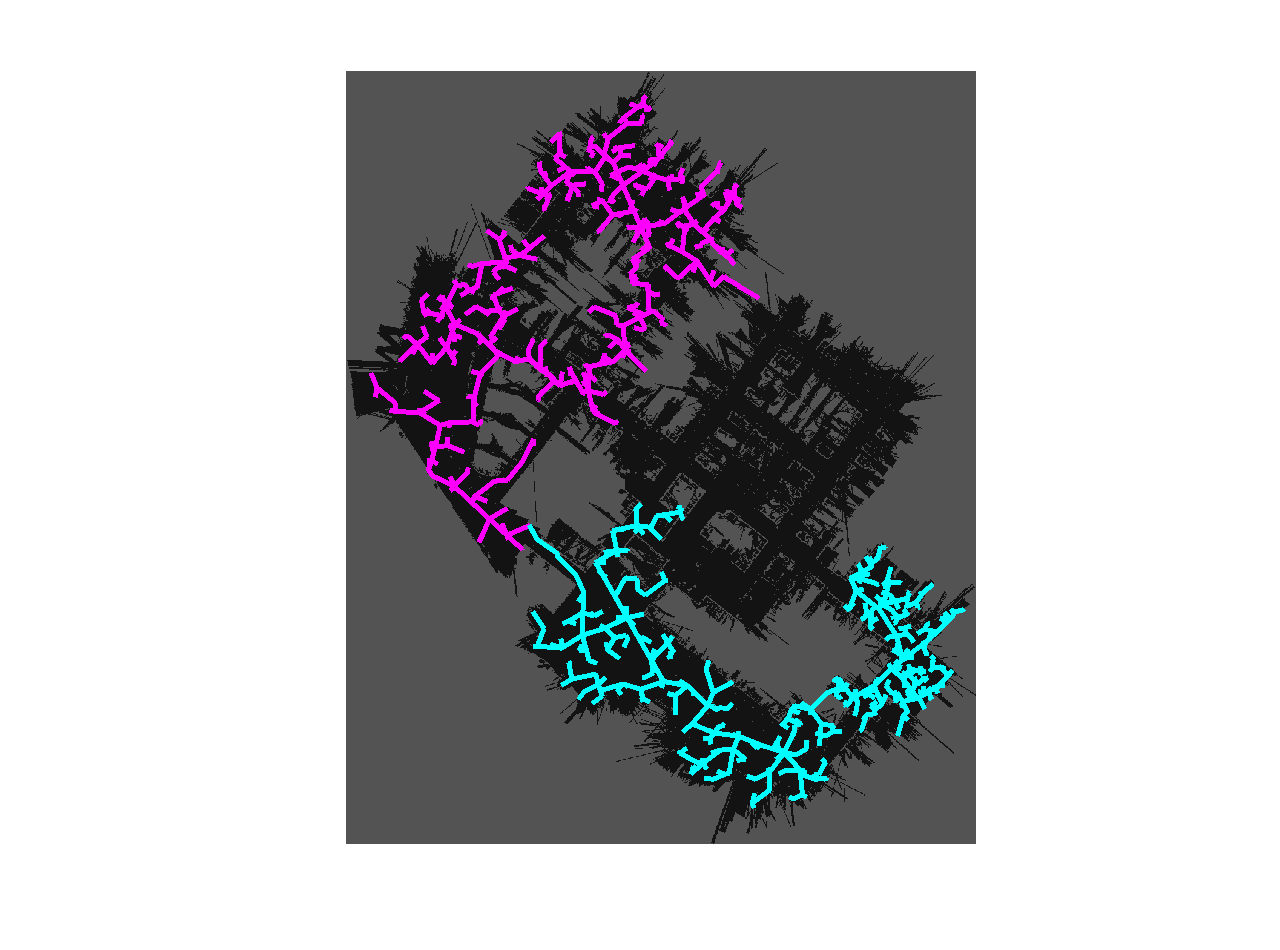}
\includegraphics[width=2.2in, trim=3.5in 1in 3in .5in, clip]{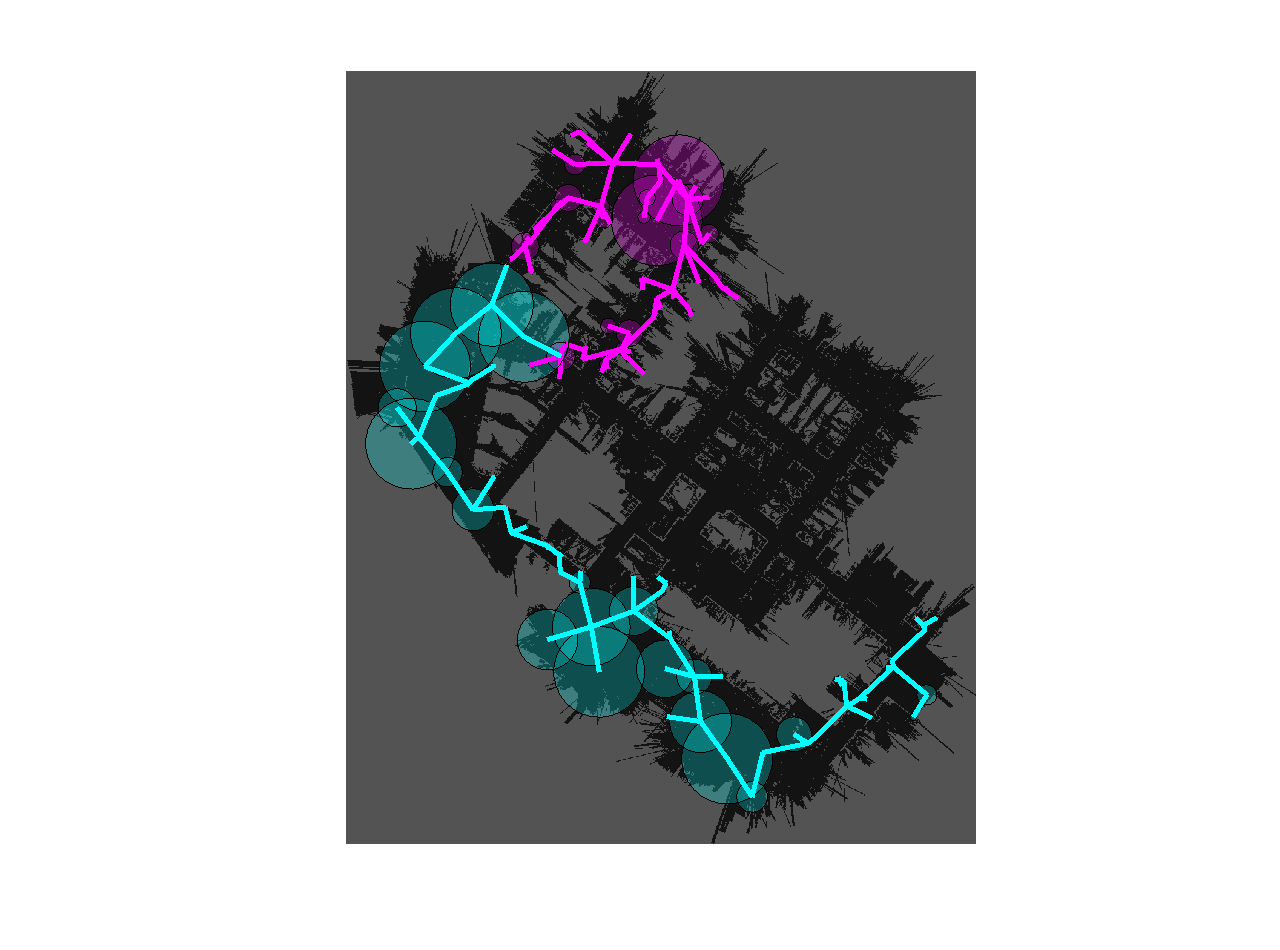}
\caption{ Left: Map of Cambridge near MIT, collected with lidar data; Middle: RRT-Connect; Right: Inexact Ball-Tree  }  \label{f:map}
 \label{f:real-map}\end{figure*}

\begin{figure*}[!ht]
\centering
\includegraphics[width=2.8in, trim=3in 1.5in 2.5in 1.5in, clip]{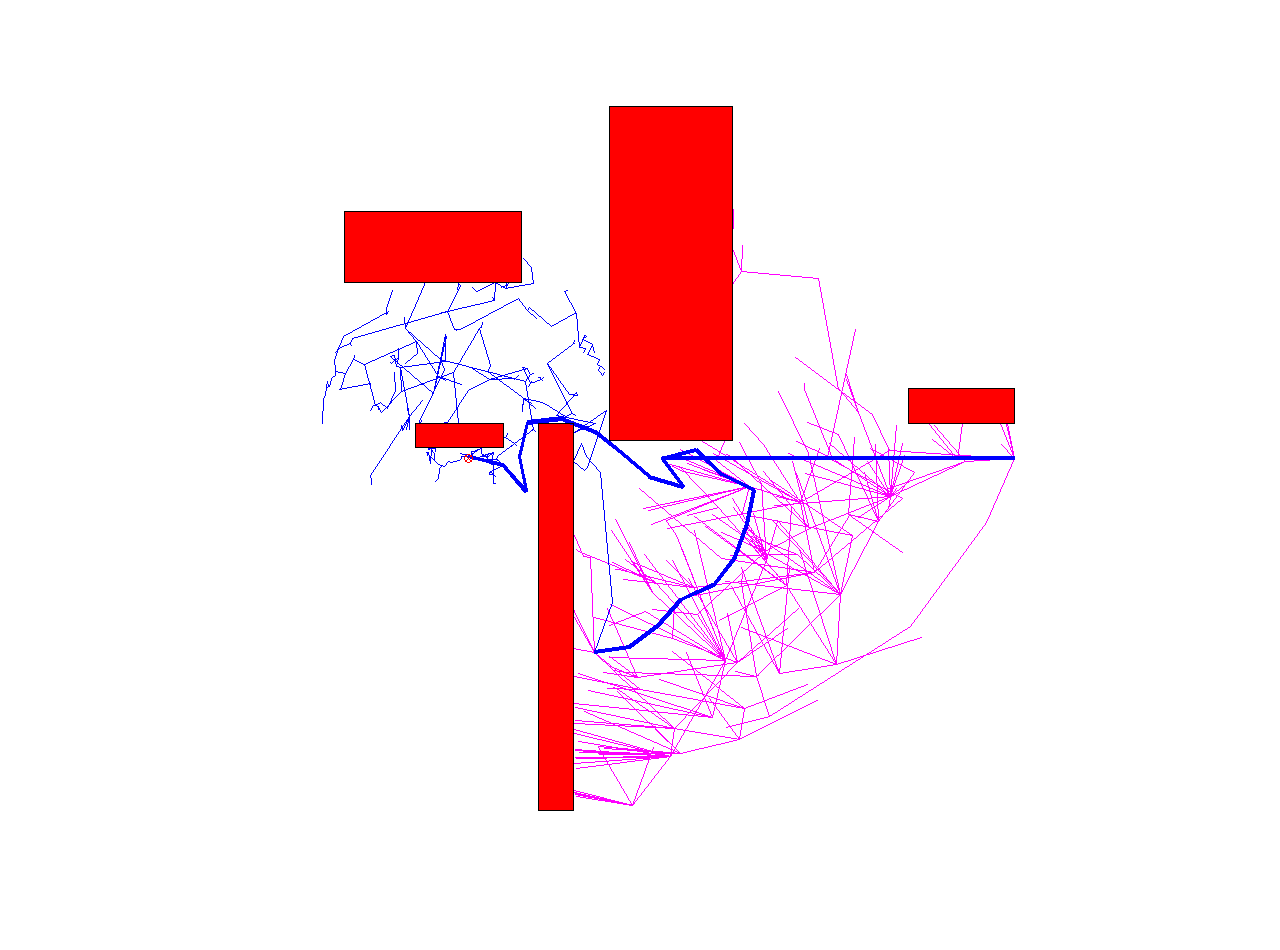} 
\hspace{.5in}
\includegraphics[width=2.8in, trim=3in 1.5in 2.5in 1.5in, clip]{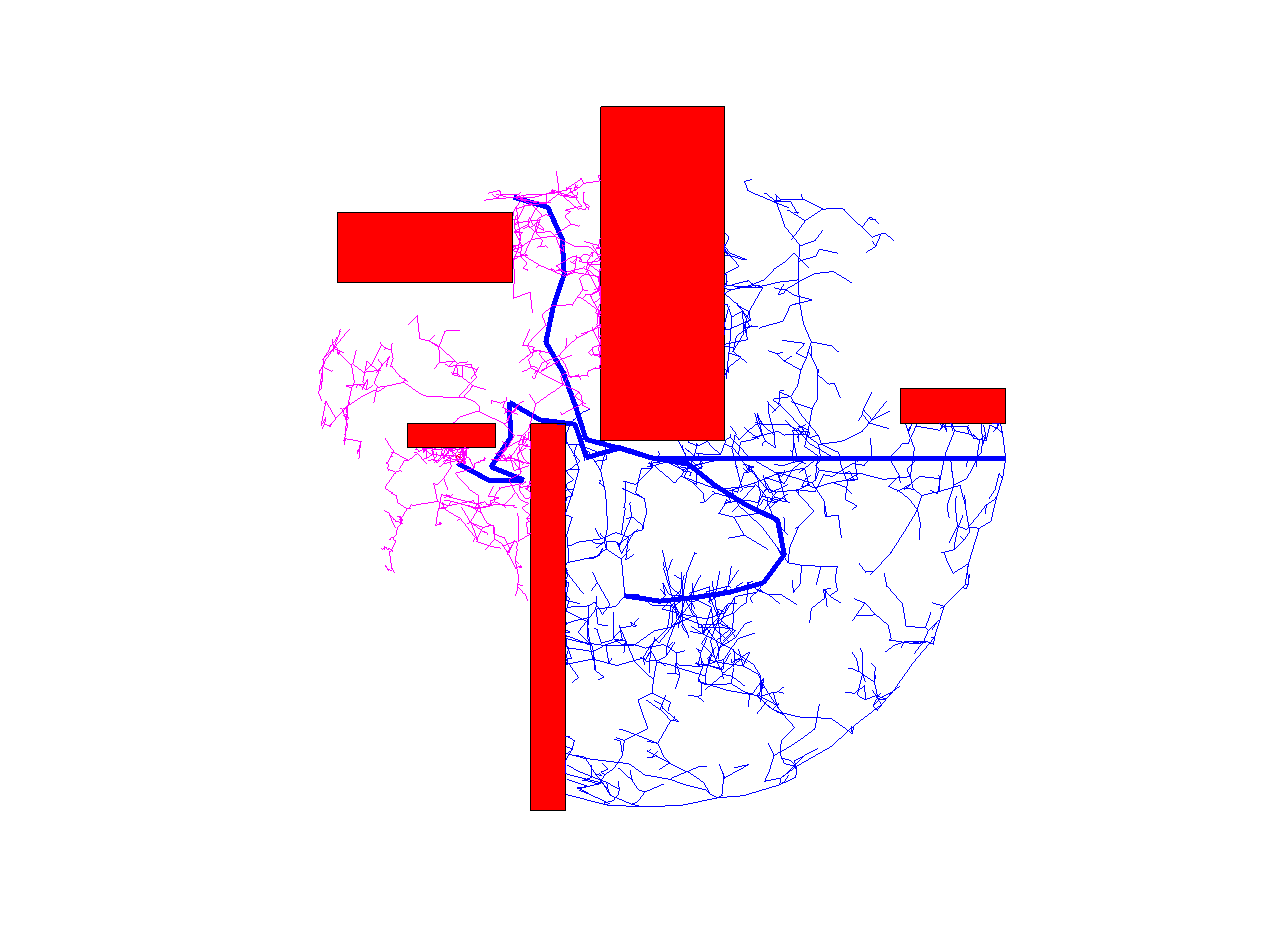}

\caption{ 10-Link arm, planning through obstacles with narrow passage. Left: Projection of Ball Tree into workspace (showing end-effector coordinate for each node in tree); Right: Projection of RRT.  }
 \label{f:10-link}
 \vspace{-10pt}
\end{figure*}

The RRT and Ball Tree algorithms were implemented in Matlab, and tested on four planning problems: (1) moving a rectangle through a bug trap [N = 3; see Figure \ref{f:bugtrap-rrt} and \ref{f:bugtrap-balls}];  (2) 10-link planar arm, tasked with moving the end point of the arm to a location in the work space while avoiding collisions [N=10; see Figure \ref{f:10-link}];  (3) Alpha Puzzle 1.0 - the most difficult of the ``Alpha'' puzzles [N=6; see bottom right of Figure \ref{f:bugtrap-rrt}]; and (4) finding a path through map data collected with Lidar mounted on a vehicle driving through Cambridge [N=2; see Figure \ref{f:real-map}].  The Ball Tree algorithm was tested against the standard RRT, which shared almost all the same code base, on a PC with a Xeon X5450 3GHZ CPU and 8.0 GB of RAM.

A comparison of RRT vs Ball Tree performance is provided in Table \ref{tab:results}. Each planner was run 30 times to collect statistics. Planning times can be dramatically improved from those presented in the table if a set of free-space samples are pre-computed. However, this pre-processing step can take several minutes (to randomly sample in the configuration space and prune out those samples which are in collision). Furthermore, for some problems such as the Alpha Puzzle, the passageways between collision surfaces can be so narrow that even with millions of pruned samples, no samples are expected to be found in the passage. Alternative sampling methods may produce better results, and have almost always been used in the past literature to solve these types of problems. In this work, however, we chose to sample uniformly from the configuration space, and include collision samples for expansion.  

\begin{table*}[hb!]
	\centering
	\caption{Comparison of run times (all data in seconds) and number of nodes required to solve several types of problems. }
	
		\begin{tabular}{ | l || c | c | c | c | c | c | c | }
			\hline
			PROBLEM & Algorithm & Mean Time (sec) & Median Time & Std Dev. & Min Time & Max Time & Mean \# Nodes \\
			\hline
			Alpha Puzzle 1.0 &  Ball-tree & 36.9 & 32.7 & 20.5 & 17.2 & 83 & 833 \\
			Alpha Puzzle 1.0 & RRT & 114.7 & 109.6 & 45.3 & 51.1 & 210.0 & 1888 \\			
			\hline
			Bug Trap & Ball-tree & 14.6 & 11.2 & 12.4 & 3.9 & 73 & 246 \\
			Bug Trap & RRT & 75.4 & 58.3 & 59.9 & 18.4 & 300+ & 17,739 \\
			\hline
			2D Lidar Data & Ball-tree & 1.1 & 1.0 & .5 & .35 & 2.3 & 156 \\
			2D Lidar Data & RRT & 2.4 & 2.1 & 1.3 & .5 & 5.9 & 663 \\						
			\hline
			10-Link Manipulator &  Ball-tree & 40.2 & 36.6 & 18.7 & 13.9 & 84.0 & 1044 \\
			10-Link Manipulator & RRT & 68.6 & 55.45 & 48.57 & 15.9 & 228.1 & 2252 \\
			\hline
		
		\end{tabular}
		\label{tab:results}
\end{table*}

The sampling approach described above, with simple straight-line expansion, worked well for each planning problem except the Alpha Puzzle 1.0, for which we could not find solutions with the RRT in reasonable time. For this problem, we used local trajectory optimization to try to find the longest collision-free path when expanding a node towards a sample. 
 With this trajectory warping approach, and using a fairly constrained sampling range, we were able to consistently solve the Alpha Puzzle in less than two minutes with a standard RRT, exploring fewer than 2000 expansions. The Ball Tree improves this performance further. The Alpha Puzzle was obtained from \cite{Yamrom}, and the collision checking was done by calling a V-COLLIDE collision detection library wrapped in a Matlab Mex function. 

A modified bug-trap type problem, shown in Figure \ref{f:bugtrap-balls}, demonstrates the potential sparsity that results from the Ball Tree.  In this problem, a rectangular robot is tasked with leaving the trap, requiring passage through a very narrow tunnel. The Ball Tree consistently required fewer than $< 2\%$ of the nodes of an RRT, which dramatically reduces the time spent checking for nearest neighbors, and also requires fewer expansion attempts, and therefore fewer collision checks to solve the problem. 

In order to test the algorithm on a real-world planning problem, we formulated a planning problem on a 2D occupancy grid representation of a real environment generated using an autonomous car (built using OctoMap 
 with a number of SICK scanners and a Velodyne).  We anticipated that spurious points in the occupancy grid could slow down our algorithm, but in practice the Inexact Ball Tree performed very well (see Figure~\ref{f:map}).
 
Finally, A 10-Link arm planning problem is shown in Figure \ref{f:10-link}. The arm must wrap through some narrow passages to find its way to the goal configuration. The figure shows examples of a few arm positions along a solution trajectory, and also displays the end-effector coordinate affiliated with each node of the tree. Although the Ball Tree finds solutions more quickly than the RRT, the gains are not as impressive as some of the other applications. This is true because in this particular problem, almost every configuration is near a collision surface in \textit{some} dimension.  Because the Ball Tree uses spherical volumes, if a single joint is near collision, then the ball radius is reduced, which penalizes the other dimensions. An obvious extension may be to use ellipses instead of balls, at the trade-off that more pruning may be required to trim the volumes to appropriate dimensions.

\section{Concluding Remarks}  \label{sec:conclusion}
\label{s:conclusion}

	\begin{figure}[t]
\centering
\includegraphics[width=2.3in, trim=4.5in 2.6in 4.5in 2.0in, clip]{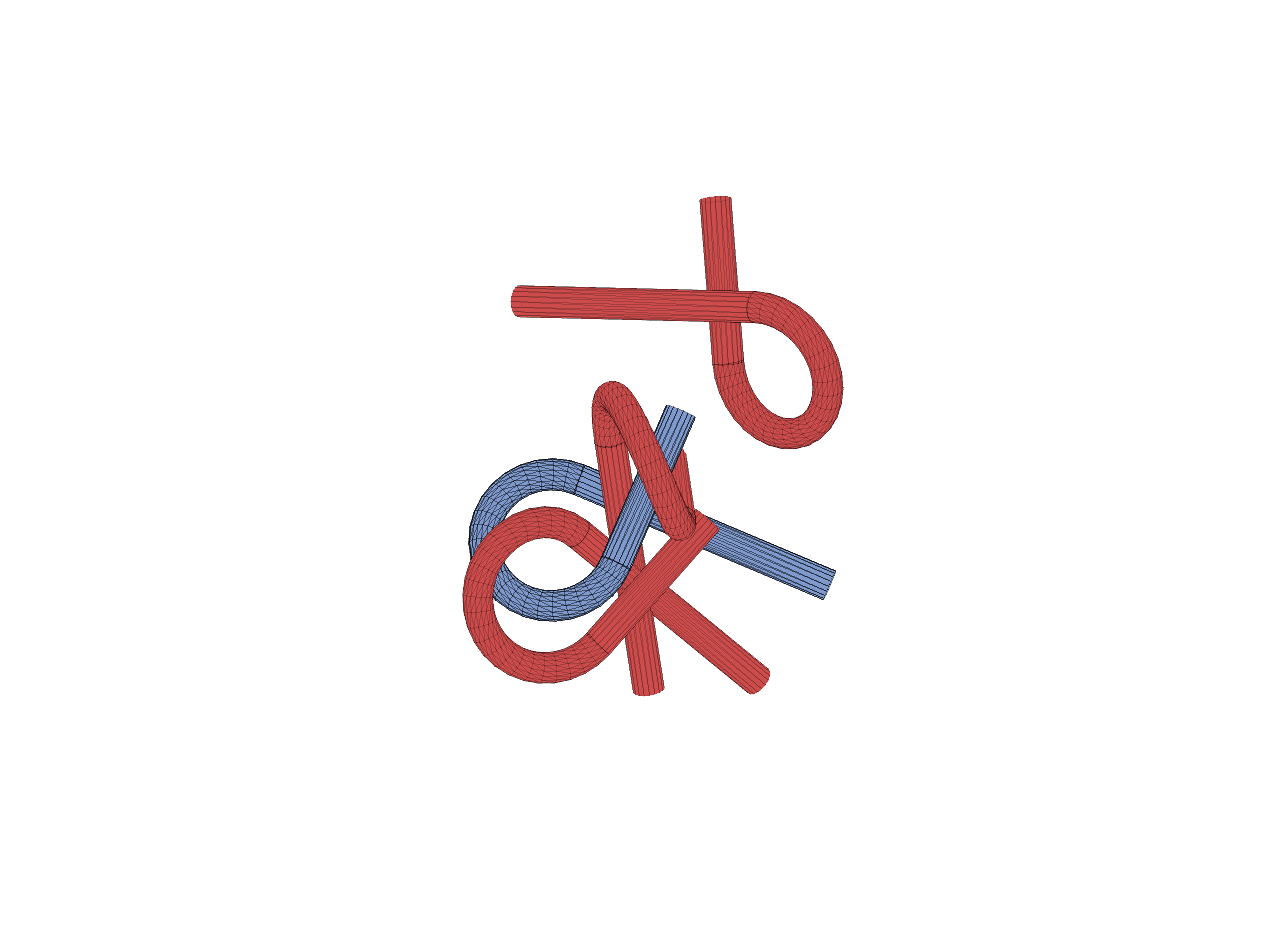} 

\caption{ Alpha Puzzle 1.0, and partial solution trajectory from Ball Tree. Puzzle has two intertwined pieces. The blue (obstacle) remains in fixed position. The red (robot) moves around the obstacle to free itself.  }
 \label{f:alpha}
\end{figure}

The algorithm presented in this paper uses volumes to characterize the free space which is reachable from a tree. The approach results in sparse trees and improved search efficiency. When applied in an RRT-like framework, a sparse tree results in fewer distance checks to compute nearest neighbors. Furthermore, fewer expansion attempts are made, so less collision checking is required.  This comes at the added computational expense of rejection sampling, but this trade off seems worthwhile for many planning applications. The bug trap type problem is particularly amenable to the Ball Tree planning method. For such problems, containing intertwined regions of large empty free space, and small narrow passages, the Ball Tree will quickly identify the empty regions, and focus on sampling near collision surfaces to try to continue expanding the tree.  In the limit where all configurations are near some obstacle, each node volume is pruned to near-zero radius, and the algorithm will converge to behave like an RRT.  

The Exact Ball Tree was presented because it is conceptually very simple, and demonstrates the rejection sampling ideas. In practice, it is difficult to ascertain the closest collision distance, so an Inexact Ball Tree method was presented, in which nodes are added to the tree with an arbitrarily large volume. The volume is verified and pruned whenever unexpected collisions are encountered.  The performance of the inexact tree actually outperforms the exact version. This is because 1) volumes are very quickly trimmed to reasonable values, typically requiring only a few expansion attempts; and 2) sharp corners in free space are more effectively covered by balls which are slightly overlapping obstacles, resulting in fewer nodes needed to cover the same space. 

Like the RRT, the Ball Tree is simple to implement, and efficiently searches for paths. It is likely that many of the modifications that work with the RRT can be directly applied to the Ball Tree as well. A number of modifications to the algorithm are possible. 
For one, the proof suggests that any distance metric can be used to construct volumes. Instead of $L_2$ hyperspheres, a metric such as $L_{\infty}$ may work better in high dimensional spaces, since the volume of a Euclidean sphere decays to zero as dimension increases. In general, volumes may be defined by sublevel-sets of any valid cost function. Additionally, instead of level-sets (in which  a volume is specified by a single number or radius), it may be interesting to use other convex volumes such as ellipses with the Euclidean distance metric, so long as there is an efficient method for computing distance to the surface of the nodes. 

	Another extension is the Multi-Ball Tree. This algorithm might start with two trees, one from the start node, and one from the goal node. If an expansion step fails, but the random sample is collision free, then a new tree is initialized with that sample as a first node.  For each sample which is not inside any tree volume, choose to expand the closest tree towards the sample. If the expansion succeeds, then check for overlap between the newly added volume, and the other trees. If there is overlap (implying a connection between trees), attempt to connect the trees. This will either trim the radius of the new node so there is no overlap between trees, or it will merge two trees together.  Continue until the tree containing the goal node is merged into the tree containing the start node.

	
	Finally, it is worth mentioning that other sampling-based approaches such as the PRM may benefit from similar ideas. The PRM lays down a number of samples in C-Space, then tries to connect proximal nodes to build a connected graph which can be searched. The process may be accomplished efficiently by iteratively sampling, and associating each sample with a volume, then checking for valid edges between overlapping volumes of the graph. The process can be repeated, building up a graph, while using rejection sampling so that only regions that are not currently in the volume of the graph are searched over.

\section*{Acknowledgment}
The authors would like to thank Michael Levashov for helping to flush out the ideas in this paper, and 
Abraham Bachrach for providing the lidar data sets of Cambridge.

\bibliographystyle{IEEEtran}
\bibliography{elib}


\end{document}